\documentclass[conference]{IEEEtran}
\usepackage{geometry}
\ifCLASSINFOpdf
\usepackage[pdftex]{graphicx}
\usepackage{epstopdf}
  \DeclareGraphicsExtensions{.pdf,.jpeg,.png,.eps}
\else
  \usepackage[dvips]{graphicx}
\fi
\usepackage{amssymb} 
\usepackage{amsfonts} 
\usepackage[cmex10]{amsmath}
\usepackage[ruled]{algorithm2e}

\usepackage{url}
\usepackage{doi}

\begin{document}
%
\title{Edge-enhancing Filters with Negative Weights}

\author{\IEEEauthorblockN{Andrew Knyazev}
\IEEEauthorblockA{Mitsubishi Electric Research Laboratories (MERL),
201 Broadway, Cambridge, MA 02139, USA\\
Email: knyazev@merl.com, 
\url{http://www.merl.com/people/knyazev}}}

\maketitle

\begin{abstract}
In 
[\doi{10.1109/ICMEW.2014.6890711}], 
a~graph-based denoising is performed by projecting the noisy image to a lower dimensional 
Krylov subspace of the graph Laplacian, constructed using nonnegative weights determined by distances between 
image data corresponding to image pixels. We~extend the construction of the graph Laplacian to the case, 
where some graph weights can be negative. 
Removing the positivity constraint provides a more accurate inference of a graph model behind the data, 
and thus can improve quality of filters for graph-based signal processing, e.g.,\ 
denoising, compared to the standard construction, without affecting the costs. 
\end{abstract}


%
\IEEEpeerreviewmaketitle

\section{Introduction}\label{s:i}
Constructing efficient signal filters is a fundamental problem
in signal processing with a vast literature; see, e.g.,\
recent papers \cite{phj12,6778405,6319316,mrs2009,ckh2014,6957555} and references there. A filter 
can be described by a transformation $F$, often non-linear, of an input signal, 
represented by a vector $x$, into a filtered signal, represented by a vector  $F(x)$.   
We~revisit some classical constructions of filters aimed at signal noise reduction, 
with the emphasis on bilateral filter, popular in image denoising \cite{1042377,4587843,Zhang2014299,M13}. 
The goal of the filter is signal smoothing, reducing a high oscillatory additive noise. 
The smoothing can be achieved by averaging, which can be 
typically interpreted as a low-pass filter, minimizing the 
contribution in the filtered signal of highly oscillatory modes, 
treated as eigevectors of a graph Laplacian;
see, e.g., \cite{Chung97a}. 

It is desirable to preserve edges in the ideal noise-free signal, 
even at the costs of an increased PSNR. 
Edge-conscious filters detect, often implicitly, the locations of the edges 
and attempt using less aggressive or anisotropic averaging at these locations. 
Fully automatic edge detection in a noisy signal is difficult, 
typically resulting in non-linear filters, i.e. where 
the filtered vector $F(x)$ depends non-linearly on the input vector $x.$ 
However, it can be assisted by a guiding signal, 
having the edges in the same locations as in the ideal signal; see,
e.g.,\ \cite{6319316,4359322,5539896}. 

Graph signal processing, introducing eigenvectors of the graph Laplacian as
natural extensions of the Fourier bases, 
sheds new light at image processing; see, e.g.,  \cite{leo_book,shuman_signal_2013,Sunil_GlobalSip13,gadde2013bilateral}.
In~\cite{TKMV14}, graph-based filtering of noisy images is performed by directly computing a projection of the image to be filtered onto a lower dimensional Krylov subspace of the normalized graph Laplacian, constructed using nonnegative graph weights determined by distances between image data corresponding to image pixels. We extend the construction of the graph Laplacian to the case, where some weights can be negative, radically departing from the traditional assumption.

\newpage
\section{Preliminaries}\label{s:p}
Let us for simplicity first assume that the guiding signal, denoted by $y$,
is available and can be used to reliably detect the locations of the edges
and, most importantly, to determine the edge-conscious \emph{linear} transformation (matrix)
$F_y$ such that the action of the filter $F(x)$ is given by the following 
matrix-vector product $F_y x = F(x).$ Having a specific construction of the guided filter 
matrix $F_y$ as a function of~$y$, one can define a self-guided non-linear filter, 
e.g.,\ as $F_x x$, which can be applied iteratively, starting with the 
input signal vector $x_0$ as follows, $x_{i+1}=F(x_i),\, i=0,1,\ldots,m$;
cf., e.g., \cite{Sarela:2005:DSS:1046920.1058110}.

Similarly, an iterative application of the linear guided filter can be used, 
mathematically equivalent to applying the powers of the square matrix $F_y$, i.e.
$x_m=\left(F_y\right)^m x_0$, thus naturally called the \emph{power method},
which is an iterative form of PCA; see, e.g.,\cite{NIPS1998_1491,Rossi2015}.
To avoid a re-normalization of the filtered signal, it is convenient to 
construct the matrix $F_y$ in the form $F_y=D_y^{-1}W_y$, where entries of the square matrix 
$W_y$ are called \emph{weighs}. The matrix $D_y$ is diagonal, made of row-sums of the 
matrix $W_y$, which are assumed to be non-zero. Thus, $D_y^{-1}W_y$ multiplied by a column-vector of ones, gives again the column-vector of ones. 

Let us further assume that the matrix $W_y$ is symmetric and that all the entries (weighs) in $W_y$ are nonnegative. 
For signal denoising, the following observations are the most important. The right eigenvector $v_1$ of the matrix $D_y^{-1}W_y$ with the eigenvalue $\mu_1=1$ is trivial, just made of ones, only affecting the signal offset. 
Since the iterative matrix $F_y=D_y^{-1}W_y$ is diagonalizable, the power method gives 
\begin{equation}
x_m=\left(F_y\right)^m x_0 = \Sigma_j \; \mu_j^m \left(v_j^T D x_0\right) v_j,
\label{eq:power}
\end{equation} 
where $1=|\mu_1|\geq|\mu_2|\geq\ldots$ are the eigenvalues of the matrix $D_y^{-1}W_y$ corresponding to the eigenvectors $v_j$ scaled such that $v_i^T D v_j=\delta_{ij}$. 
The power method, according to~\eqref{eq:power}, 
suppresses contributions of the eigenvectors corresponding to
the smallest eigenvalues. Thus, the matrix $W_y$ needs to be constructed in such a way 
that these eigenvectors represent the noisy part of the input signal, while 
the other eigenvectors are edge-conscious; cf. anisotropic diffusion \cite{PM87, PM90, GraphSP_HeatKernel:2008}.

Let us introduce the guiding Laplacian $L_y=D_y-W_y$ and 
normalized  Laplacian $D_y^{-1}L_y=I-D_y^{-1}W_y$ matrices. 
In~\cite{TKMV14}, the power method \eqref{eq:power} is replaced with a projection of the image vector $x$ 
to be denoised onto a lower dimensional Krylov subspace of the guiding normalized graph Laplacian $D_y^{-1}L_y$ and implemented, e.g., using the Conjugate Gradient (CG) method; see, e.g., \cite{Hestenes&Stiefel:1952,G97,823968}. 

\newpage
\section{Motivation} \label{s:m}
One of the most popular edge-preserving denoising filters is the bilateral filter (BF), see, e.g.,\ \cite{TM98, PKTD09} and references there, which takes the weighted average of the nearby pixels.
The weights  $w_{ij}$ may depend on spatial distances and signal data similarity, e.g.,\
\begin{equation}\label{eq2}
 w_{ij}=\exp\left(-\frac{\left\|p_i-p_j\right\|^2}{2\sigma_d^2}\right)
 \exp\left(-\frac{\left\|y[i]-y[j]\right\|^2}{2\sigma_r^2}\right),
\end{equation}
where $p_i$ denotes the position of the pixel $i$, the value $y[i]$ is the signal intensity, and $\sigma_d$ and $\sigma_r$ are filter parameters.
To simplify the presentation and our arguments, we further assume that 
the signal is scalar on a one-dimensional uniform grid, setting 
without loss of generality the first multiplier in \eqref{eq2} to be $1$, 
and that the weights $w_{ij}$ are computed only for the nearest neighbors and set to zero otherwise.

\begin{figure}[!t]
\centering
\includegraphics[width=\linewidth]{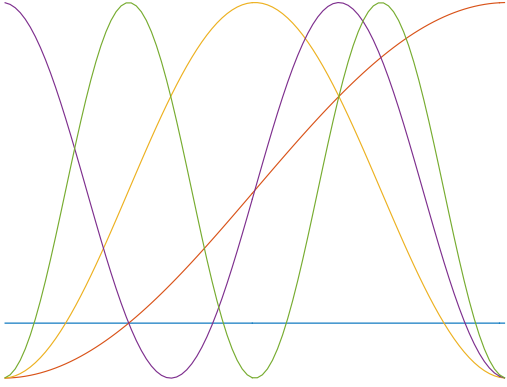}
\caption{Discrete cosine transform low frequency modes.}
\label{fig:1}
\end{figure}

Let us start with a constant signal, where $y[i]-y[j]=0$. 
Then, $w_{i-1\, i}=w_{i\, i}= w_{i\, i+1}=1$ and the graph
Laplacian $L_y=D_y-W_y$ is a tridiagonal matrix that has nonzero entries 
$1$ and $-1$ in the first row, $-1$ and $1$ in the last row, and 
$[-1\,\; 2\, -1]$ in every other row. This is a standard three-point-stencil finite-difference 
approximation of the negative second derivative of functions 
with homogeneous Neumann boundary conditions, i.e., vanishing 
first derivatives at the end points of the interval. Its eigenvectors 
are the basis vectors of the discrete cosine transform; see the first five 
low frequency eigenmodes (the eigenvectors corresponding to the smallest eigenvalues) of  $L_y$
in Figure \ref{fig:1}.  
As can be seen in Figure~\ref{fig:1}, all smooth low frequency eigenmodes 
turn flat at the end points of the interval, due to the Neumann conditions. 

The key observation is that
the Laplacian row sums in the first and last rows vanish for \emph{any} signal,
according to the standard construction of the graph Laplacian,
no matter what formulas for the weights are being used!
Thus, \emph{any} low pass filter 
based on low frequency eigenmodes of the graph Laplacian flattens 
the signal at the end points. 

\begin{figure}[!t]
\centering
\includegraphics[width=\linewidth]{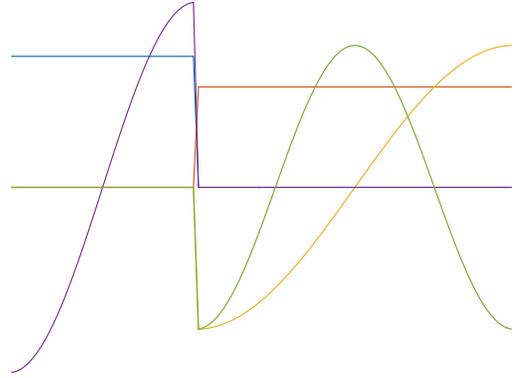}
\caption{Edge-preserving low frequency eigenmodes.}
\label{fig:2}
\end{figure}

Let us now use formula \eqref{eq2} for a piece-wise constant
guiding signal $y$ with the jump large enough to result in a small value  
$w_{i\, i+1}=w_{i+1 \, i}$ for some index~$i$. The first five vectors of the 
corresponding Laplacian are shown in Figure \ref{fig:2}.  
All the plotted in Figure \ref{fig:2} 
vectors are aware of the jump, representing an edge in our 
one-dimensional signal~$y$, but they are also all flat on both sides of the edge! 
Such a flatness is expected to appear for any guiding signal $y$ giving 
a small value $w_{i\, i+1}=w_{i+1 \, i}$.

The presence of the flatness in the low frequency modes of the graph Laplacian $L_y$ on both sides of the edge in the guiding signal $y$ is easy to explain. When the value $w_{i\, i+1}=w_{i+1 \, i}$ is small relative to other entries, the matrix  $L_y$ becomes 
nearly block diagonal, with two blocks, which approximate graph Laplacian matrices 
of the signal $y$ restricted to sub-intervals of the signal domain 
to the left and to the right of the edge. 

The low frequency eigenmodes
of the graph Laplacian $L_y$ approximate combinations of the 
low frequency eigenmodes of the graph Laplacians on the sub-intervals.
But each of the low frequency eigenmodes of the graph Laplacian on the sub-interval
suffers from the flattening effect on both ends of the sub-interval, as explained above. 
Combined, it results in the flatness in the low frequency modes of the graph Laplacian $L_y$
on both sides of the edge.
For denoising, the flatness of the vectors determining the low-pass filter 
may have a negative effect for self-guided denoising even of piece-wise constant signals,
if the noise is large enough relative to the jump in the signal, as shown in Section \ref{s:eef}.

The attentive reader notices that method~\eqref{eq:power} is based on $D_y^{-1}W_y$, related to the \emph{normalized} graph Laplacian $D_y^{-1}L_y$, not the Laplacian $L_y$ used in our arguments above. Although the diagonal matrix
$D_y$ is not a scalar identity, and so the eigenvectors of $D_y^{-1}L_y$,
not plotted here, and of $L_y$ are different, the difference is not qualitative enough to noticeably change the figures and invalidate our explanation. 
 
\section{Negative weights in spectral graph partitioning and for signal edge enhancing}
\label{s:n}
\begin{figure}[!t]
\centering
\includegraphics[width=\linewidth]{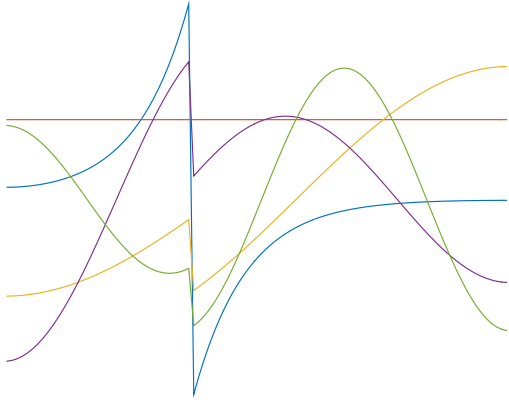}
\caption{Edge-enhancing low frequency eigenmodes, small negative.}
\label{fig:3}
\end{figure}
The low frequency eigenmodes of the graph Laplacian play a fundamental role 
in spectral graph partitioning, which is one of the most popular 
tools for data clustering; see, e.g.,\ \cite{Shi:2000:NCI:351581.351611,ng2002spectral,k2003}.
A limitation of the conventional spectral clustering approach is embedded in its definition based on the weights of graph, which must be nonnegative, e.g.,\ based on a distance measuring relative similarities of each pair of points in the dataset. For the dataset representing values of a signal, e.g.,\ pixel values of an image, formula \eqref{eq2} is a typical example of 
determining the nonnegative weights, leading to the graph adjacency matrix $W_y$ with nonnegative entries, as assumed in Section \ref{s:p} and in all existing literature. 

In applications, data points may represent feature vectors or functions, allowing the use of correlation for their pairwise comparison. The correlation can be negative, or,  more generally, points in the dataset can be dissimilar, contrasting each other.
In conventional spectral clustering, the only available possibility to handle such a case is to replace the anticorrelation, i.e. negative correlation, of the data points with the uncorrelation, i.e. zero correlation. The replacement changes the corresponding negative entry in the graph adjacency matrix to zero, to enable the conventional spectral clustering to proceed, but nullifies a valid comparison. 

A common motivation of spectral clustering comes from analyzing a mechanical vibration model in a spring-mass system, where the masses that are tightly connected have a tendency to move synchronically in low-frequency free vibrations; e.g.,~\cite{Park20143245}. Analyzing the signs of the components corresponding to different masses of the low-frequency vibration modes of the system allows one to determine the clusters. The mechanical vibration model may describe conventional clustering when all the springs are pre-tensed to create an attracting force between the masses. However, one can pre-tense some of the springs to create repulsive forces!

\begin{figure}[!t]
\centering
\includegraphics[width=\linewidth]{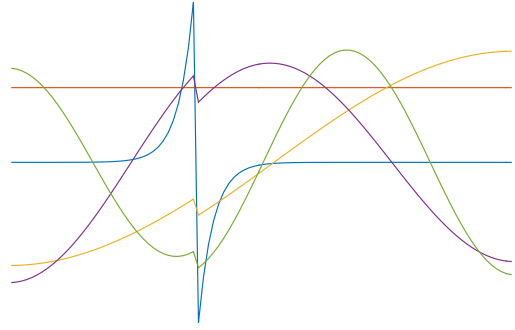}
\caption{Edge-enhancing low frequency eigenmodes, more negative.}
\label{fig:4}
\end{figure}
In the context of data clustering formulated as graph partitioning, that corresponds to negative entries in the adjacency matrix. The negative entries in the adjacency matrix are not allowed in conventional graph spectral clustering. Nevertheless, the model of mechanical vibrations of the spring-mass system with repulsive springs remains valid, motivating us to consider 
the effects of negative graph weights. 

In the spring-mass system, the masses, which are attracted, would move together synchronically in the same direction in low-frequency free vibrations, while the masses, which are repulsed, have the tendency to move synchronically in the opposite direction. 
Using negative, rather than zero, weights at the edge of the guiding signal $y$ 
for the purposes of the low-pass filters thus 
is expected to repulse the flatness of low frequency eigenmodes
of the graph Laplacian $L_y$ on the opposite sides of the edge of the signal $y$, 
making the low frequency eigenmodes to be edge-enhancing, rather than just edge-preserving; 
cf. \cite{Durand:2002:FBF:566654.566574} on sharpening. 

Figures \ref{fig:3} and \ref{fig:4} demonstrate the effect of edge-enhancing, as a proof of concept.
Both Figures \ref{fig:3} and \ref{fig:4} display the five eigenvectors for the five smallest eigenvalues of the same tridiagonal graph Laplacian as that corresponding to Figure \ref{fig:2}
except that the small positive entry of the weights $w_{i\, i+1}=w_{i+1 \, i}$ for the same $i$  
is substituted by $-0.05$ in Figure \ref{fig:3} and by $-0.2$ in Figure \ref{fig:4}. 
The previously flat around the edge eigenmodes in Figure \ref{fig:2}  
are repelled in opposite directions on the opposite sides of the edge in Figures \ref{fig:3} and~\ref{fig:4}. 

Negative weights require caution, since even small changes 
dramatically alter the behaviors of the low frequency eigenmodes around the 
edge, as seen in Figures \ref{fig:3} and \ref{fig:4}. Making the 
negative value more negative, we observe by comparing Figure \ref{fig:3} to Figure \ref{fig:4}
that the leading eigenmode, displayed using the blue color in both figures, corresponding to the smallest nonzero eigenvalue forms a narrowing layer around the 
signal edge, while other eigenmodes become less affected by the change
in the negative value. 

\section{Edge-enhancing filters}\label{s:eef}
\begin{figure}[!t]
\centering
\includegraphics[width=\linewidth]{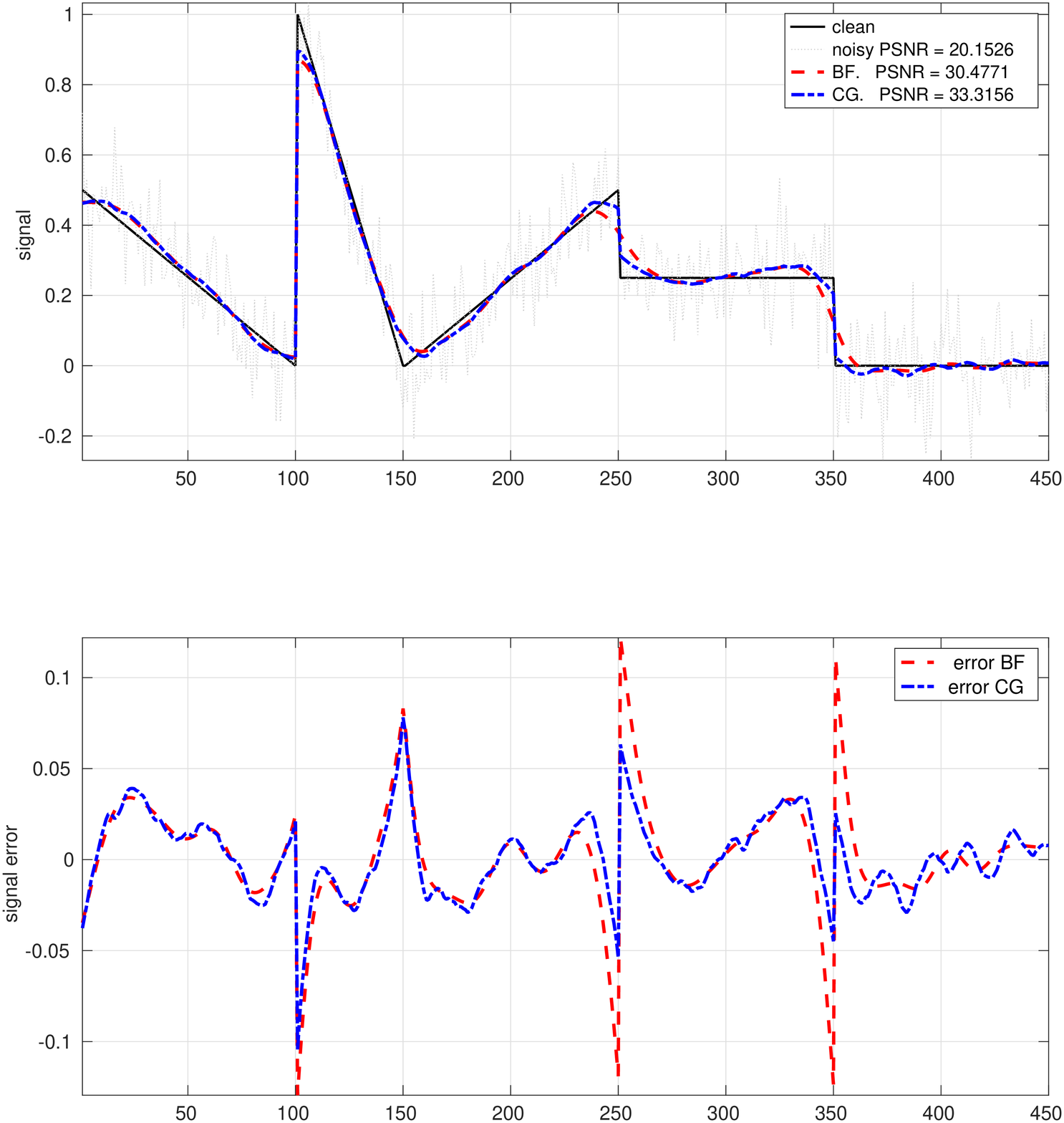}
\caption{Edge-preserving filtering.}
\label{fig:zero}
\end{figure}

In this section, as a proof of concept, we numerically test the proposed edge-enhancing filters
on a toy one-dimensional example using the classical nonlinear self-guiding BF and 
a guided (by a noiseless signal) BF
accelerated with a conjugate gradient (CG-BF) method, as suggested in \cite{TKMV14}. 
The specific CG algorithm used in our tests is as described in Algorithm~\ref{alg:CG}.

\begin{algorithm}
{Input: signal vector to be filtered $x_0$, matrices $D_y$ and $L_y$} \\
{$r_0=-L_y x_0$}\\
    \For{$k=0,1,\ldots,m-1$}{
    $s_k = D_y^{-1} r_k$\\
    \eIf{$k=0$}{
      $p_0 = s_0$\\
    }
    {
    $p_k = s_k + \beta_k p_{k-1}$, where \\
    $\displaystyle \beta_k = \frac{(s_k,r_k)}{(s_{k-1},r_{k-1})}$ 
    }
    $q_k=L_y p_k$\\
    $\displaystyle \alpha_k = \frac{(s_k,r_k)}{(p_k,q_k)}$\\
    $x_{k+1} = x_k + \alpha_k p_k$\\
    $r_{k+1} = r_k - \alpha_k q_k$\\
    }
    {Output: filtered vector $x_m$}
    \caption{Conjugate Gradient Guided Filter}
    \label{alg:CG}
\end{algorithm}

\begin{figure}[!t]
\centering
\includegraphics[width=\linewidth]{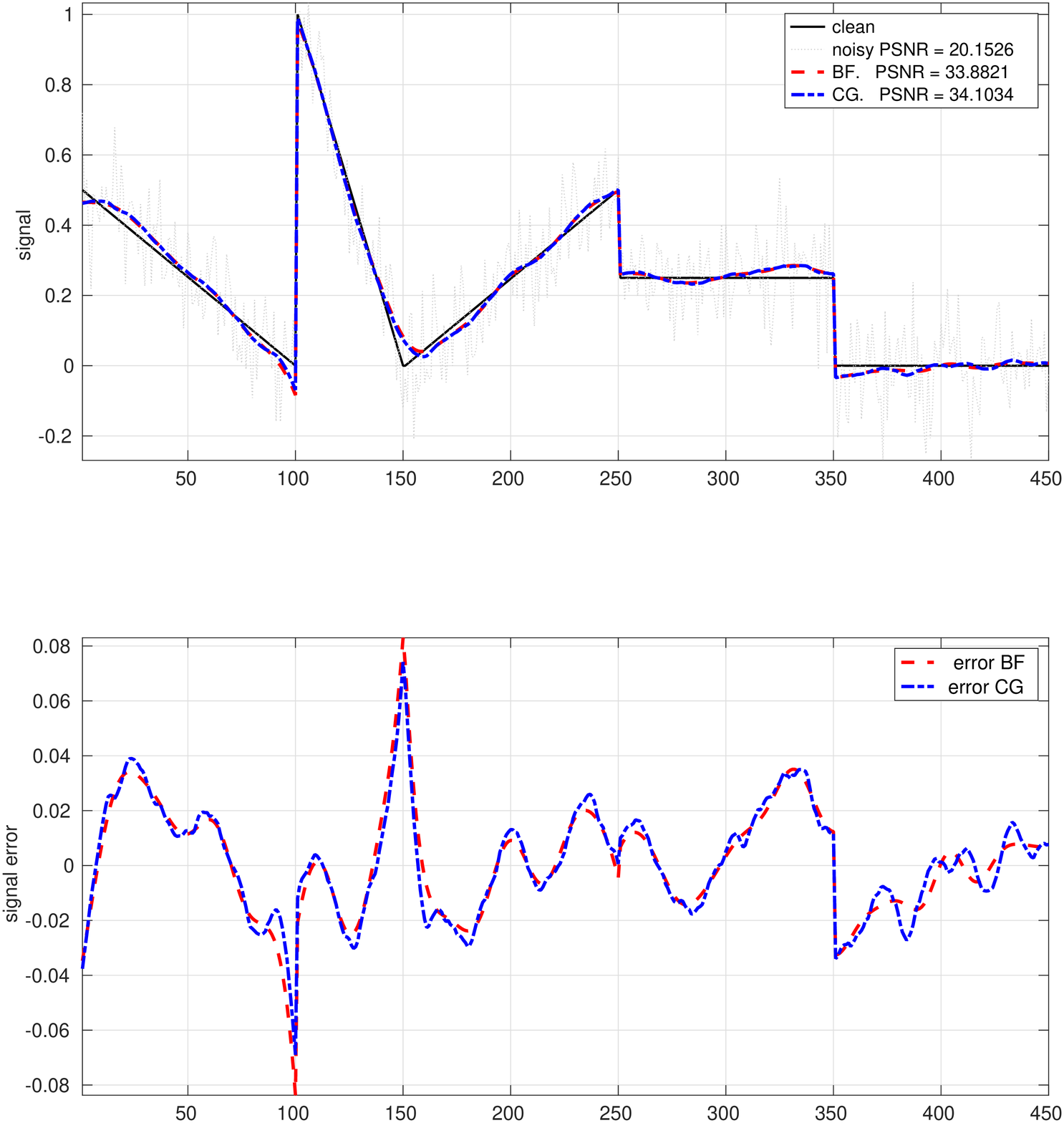}
\caption{Edge-enhancing filtering.}
\label{fig:negative}
\end{figure}
The noise is additive Gaussian, and the noisy signal is displayed using grey dots.
The nonzero weights are computed by \eqref{eq2} with $\sigma_d=0.5$ and $\sigma_1=0.1$ 
only for $j=i-1, \, i, \, i+1$, resulting in tridiagonal matrices $W$ and $L$. 
BF is self-guided, with  $W$ and $L$ recomputed on every iteration
using the current approximation $x_k$ to the final filtered signal $x_m$.
CG-BF uses the fixed  nonzero weights computed also by \eqref{eq2}, but for the 
noiseless signal $y$ resulting in the fixed tridiagonal matrices $W_y$ and $L_y$. 
The number of iterations in BF, 100, and CG-BF, 15, is tuned to match the errors.
Formula \eqref{eq2} puts ones on the main diagonal of $W$, thus for 
small positive or negative $w_{i\, i+1}=w_{i+1 \, i}$ the matrix $D$ is well-conditioned. 

Figure \ref {fig:zero} demonstrates the traditional approach with nonnegative weighs.
We observe, as discussed in Section~\ref{s:m}, flattening at the end points. 
Moreover, there is noticeable edge smoothing in all corners, 
due to a large noise and relatively small number of signal samples, despite of the use of the edge-preserving 
formula \eqref{eq2}. 
We set tuned negative graph weights $-2\times 10^{-3}$, $-10^{-3}$, $-10^{-8}$ 
for $i=100,\, 250$, and $350$ correspondingly, 
to obtain 
Figure \ref{fig:negative}, which shows dramatic improvements both in terms of PSNR 
and edge matching, compared to  Figure \ref {fig:zero}.

\section{Conclusion}
The proposed novel technology of negative graph weights allows 
designing edge enhancing filters, as explained theoretically and
shown numerically for a synthetic example. 
Our future work concerns determining the optimal negative weights, testing the concept 
for image filtering, and exploring its advantages in spectral data clustering using correlations. 

\newpage
\IEEEtriggeratref{17}


%


\bibliographystyle{IEEEbib}
\bibliography{refs}  

{\let\thefootnote\relax\footnote{Accepted to IEEE GlobalSIP Conference 2015.}}

\end{document}